# Advance Fake Video Detection via Vision Transformers

Joy Battocchio
joy.battocchio@unitn.it
University of Trento
Italy

Stefano Dell'Anna
stefano.dellanna@unitn.it
University of Trento
Italy

Andrea Montibeller
andrea.montibeller@unitn.it
University of Trento
Italy

Giulia Boato
giulia@truebees.eu
University of Trento and Truebees srl
Italy

## ABSTRACT

Recent advancements in AI-based multimedia generation have enabled the creation of hyper-realistic images and videos, raising concerns about their potential use in spreading misinformation. The widespread accessibility of generative techniques, which allow for the production of fake multimedia from prompts or existing media, along with their continuous refinement, underscores the urgent need for highly accurate and generalizable AI-generated media detection methods, underlined also by new regulations like the European Digital AI Act. In this paper, we draw inspiration from Vision Transformer (ViT)-based fake image detection and extend this idea to video. We propose an original framework that effectively integrates ViT embeddings over time to enhance detection performance. Our method shows promising accuracy, generalization, and few-shot learning capabilities across a new, large and diverse dataset of videos generated using five open source generative techniques from the state-of-the-art, as well as a separate dataset containing videos produced by proprietary generative methods.

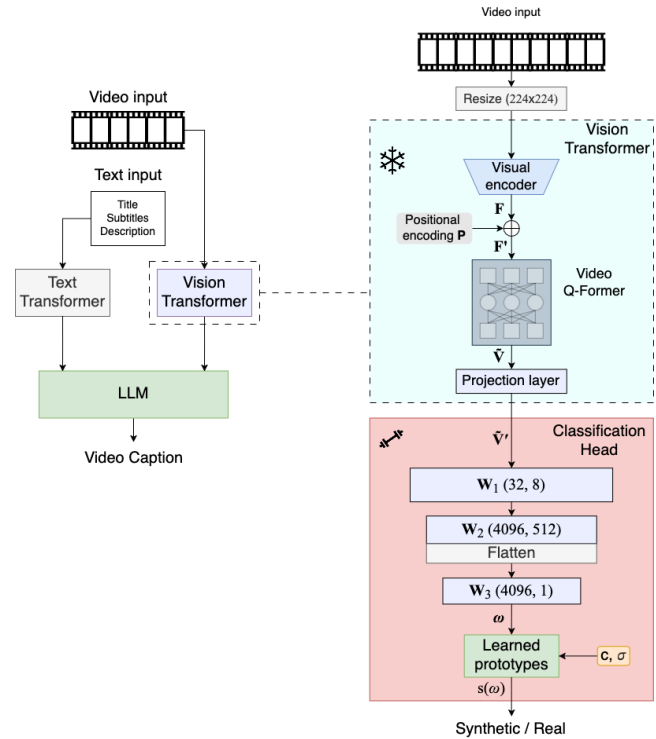

Figure 1: Overview of the proposed solution based on the Vision Transformer (ViT) from [11], originally trained for video captioning. In our framework, the ViT weights remain frozen, enabling the extraction of temporally aware, high-level semantic features from video frames. A classification head, composed of three linear layers—$W_1$, $W_2$, and $W_3$—with kernel sizes specified in the figure, is trained to further refine the ViT features $\tilde{V}'$ and distinguish real video features from synthetic ones, by using our implementation of the learned prototypes proposed in [31]. The proposed solution is trained on VideoDiffusion, a new, large, and diverse benchmarking dataset introduced in this paper.

## 1 INTRODUCTION

The rapid advancements in AI-based generative multimedia techniques have led to the creation of hyper-realistic content, making it increasingly challenging to distinguish between real and synthetic media [2]. This issue becomes particularly concerning when such media is used to spread misinformation, propaganda, fake news, or when people are unaware of interacting with AI-generated media, as also stated in the European Digital AI Act for transparent AI [36].

A wide range of generative methods exists in the literature for synthetic image creation. Generative Adversarial Networks (GANs) [19, 25] can synthesize from noise highly realistic images, which can easily fool our human visual system and be perceived as real [30, 35]. The recent advent of Diffusion Models (DMs) has further accelerated the democratization of multimedia generation, by producing hyper-realistic images from textual prompts via the *prompt-to-image* generative pipeline [38].

Recent advancements in DM-based image generation have rapidly extended to video generation [9, 12, 26, 28], achieving remarkable spatio-temporal coherence within a short time frame [28]. As a result, distinguishing DM-generated videos from real footage has become increasingly difficult. Furthermore, beyond the classic *prompt-to-media* pipeline, new approaches such as *media-to-media* and *media, prompt-to-media* [18, 26] generation have emerged, where *media* refers to an image, frame, or video.

*Prompt-to-video* models [9, 28] often combine a 3D Variational Autoencoder (VAE) with a Large Language Model (LLM)-based text encoder utilizing causal attention. *Video,prompt-to-video* methods [18, 26] can modify existing videos using noise shuffling for temporal consistency, sometimes employing Denoising Diffusion Implicit Models (DDIM) inversion [23] to retain structural integrity. *Frame, prompt-to-video* approaches may include (i) random-mask diffusion for smooth scene transitions [12, 50] and (ii) source frame propagation for video edits [18]. Despite their common reliance on DMs, these architectural differences pose challenges for detection methods generalization.

While forensic techniques for detecting GAN- and DM-generated images are well-studied [6, 13, 14], research on DM-generated video detection is still in its early stages. Existing approaches are primarily frame-based and rely on fake image detection frameworks, employing Convolutional Neural Networks (CNNs) [45] or Vision Transformer (ViT) [7] embeddings. Only recently, new methods begun to combine and exploit inconsistencies at both frame level and dense optical flow level [44] for DM-generated video detection [3]. In contrast, spatio-temporal models, such as 3D CNNs like ResNet3D, have been also proposed [40], but their effectiveness on DM-generated videos remains underexplored.

In this paper, we specifically address the problem of detecting synthetic videos generated by DMs. Inspired by very recent approaches [3, 15], we propose a novel detection method, see Fig. 1, leveraging the ViT model from [11], a method for video prompt generation. Specifically, similar to [7, 15], we freeze the ViT parameters and utilize embeddings extracted from video frames, performing the detection with a lightweight classification head. However, unlike prior work, our method processes sequences of $J$ consecutive frames rather than analyzing frames independently, exploiting the available temporal information.

Furthermore, we introduce VideoDiffusion, a new large-scale and diverse dataset containing more than ten thousand videos generated using *frame,prompt-to-video* and *video,prompt-to-video* pipelines, following the bias-free generation paradigm proposed in [22].

We compare our proposed solution with state-of-the-art (SoA) methods across multiple scenarios. First, we train both our solution and the SoA methods on two classes of uncompressed video frames from VideoDiffusion and test them on video frames from the whole dataset, both subject to and not subject to H.264 compression. Thus, we evaluate how well our solution and the SoA can generalize on a disjoint dataset of fully synthetic videos generated by proprietary models like SORA [9], LUMA-AI [29], and RunwayML [33] using the *prompt-to-video* pipelines.

Our results show that, while comparable results are obtained by all methods on uncompressed video frames, our approach outperforms existing SoA solutions in terms of single-class True Positive and True Negative Rates, and Area under the Curve for H.264 compressed videos, and generalization to unseen techniques and generative pipelines. In the end, similarly to [15], we highlight the encouraging few-shot learning capabilities of the proposed ViT-based model for fake video detection.

The remainder of this paper is organized as follows: Section 2 reviews related works, while Section 3 details the new VideoDiffusion dataset. Section 4 describes our detection method, and Section 5 presents various experimental results. Finally, we conclude and discuss the work presented in Section 6.

## 2 RELATED WORKS

To prevent the spread of fake images, the multimedia forensics literature offers various methods based on CNNs [13, 20] as well as multimodal features [15]. For instance, ResNet-50 [13] has been frequently used and adapted for fake image detection. These adjustments consisted in modifying the architecture to remove downsampling in the first layer [20], enhancing data processing during training [13], and adopting a contrastive learning paradigm to increase accuracy [20]. Similarly, Bayar et al. [5] proposed a constrained CNN architecture designed for image manipulation detection, allowing the adaptive learning of manipulation features to accurately identify the type of editing applied to an image. More recently, Cozzolino et al. [15] explored the use of pre-trained vision-language models (VLMs), such as CLIP, for detecting synthetic media. In their work, Cozzolino et al. [15] developed a lightweight detection strategy utilizing CLIP ViT features and training a simple classifier on a limited set of semantically aligned real and synthetic images. This solution [15] achieved SoA performance in challenging scenarios involving various image generative models, including DMs.

As mentioned in the Introduction, while numerous methodologies have been proposed for detecting DM-generated images, far fewer have been developed for DM-generated videos. This is due to the novelty of DM-based generative models for videos and the limited availability of datasets containing DM-generated videos [3, 45].

Vahdati et al. [45] extended MISLNet [5], a constrained CNN architecture originally designed for image manipulation detection, to detect fake videos. They achieved this by training MISLNet on H.264-compressed DM-generated video frames and aggregating single-frame prediction scores through summation before applying the softmax function. Similarly, they demonstrated that other CNN architectures initially developed for fake image detection, such as ResNet-50 [13], can be extended to fake video detection.

Bohacek et al. [7] extended the CLIP-based fake image detector proposed in [15], highlighting the advantages of leveraging CLIP ViT embeddings to capture semantic inconsistencies in fake multimedia. Specifically, they developed task-specific CLIP embeddings (FT-CLIP) and an unsupervised forensic technique that requires no explicit training, enhancing generalization across different content types and synthesis models while extending the analysis from images to videos.

Bai et al. [3] proposed AIGVDet, a two-branch architecture in which two ResNet-50 models are trained separately on RGB video frames and dense optical flows extracted using RAFT [44], a pretrained deep neural network for dense optical flow estimation. This approach exploits both spatial and temporal inconsistencies affecting synthetic videos. The extracted features are then processed by separate classifiers, and the classification scores are fused using a weighted sum, with the final video classification determined by averaging frame-based results.

Nevertheless, the solutions proposed in [45], [7], and [3] still rely on single-frame or frame-pair (for the estimation of the optical flow) predictions. These predictions are subsequently aggregated through summation, majority voting, or averaging, but hardly generalize across different generative pipeline, as we will demonstrate in our experiments.

To exploit both spatial and temporal inconsistencies present in individual and consecutive frames, Roy et al. [40] investigated various methods based on 3D-CNNs, including ResNet3D, demonstrating the advantages of using a 3D architecture over a 2D approach. However, they only focused on videos from the FaceForensics++ dataset [42].

To generalize across multiple generative pipelines and video compression while leveraging spatio-temporal features left by video generative techniques, we introduce a new solution based on the ViT proposed in [11] for video captioning, as depicted in Fig. 1. In contrast to previous ViT-based fake media detectors [7, 15], the selected ViT extracts high-level spatio-temporal semantic features from batches of $J$ frames. These are then further refined by a shallow, three-layer head and used for the classification by implementing the so-called learned prototypes [31], which allows to better model the real video class distribution. As in [7, 15], the ViT weights remain frozen, and only the classification head is trained (more details in Sect. 4). The proposed solution shows superior robustness to unknown generative pipelines and compression, as well as few-shot learning capabilities, as will be discussed in Sect. 5.

Moreover, to address the scarcity of datasets containing videos generated by SoA DM-based techniques, we present here a novel, large, and diverse benchmarking dataset called VideoDiffusion.

## 3 VIDEODIFFUSION DATASET

In this paper, we introduce VideoDiffusion,[1] a novel and extensive dataset comprising over ten thousand videos generated using five different DM-based video generative techniques [12, 18, 23, 26, 50], with various levels of compression. To construct our dataset, we drew inspiration from the bias-free methodology proposed in [22], in which synthetic media are generated starting from real data, through *media, prompt-to-media* pipelines. This allowed us to create a semantically aligned dataset pairing real media with their AI-generated counterparts, thus ensuring that any observed differences stem from subtle artifacts introduced during the generation process.

[1]The dataset and prompts are available at https://github.com/MMLab-unitn/VideoDiffusion-IHMMSec25

Table 1: Number of original video sequences from different sources, with their video resolution and frame rate.

| Dataset | Original resolution | | | Total |
| --- | --- | --- | --- | --- |
| | 3840x2160 | 1920x1080 | 1280x720 | |
| Youtube-8m [1] | - | 33 | 91 | 124 |
| Sports-1M [24] | - | - | 105 | 105 |
| Vision [43] | - | 96 | - | 96 |
| FloreView [4] | 1 | 127 | 34 | 162 |
| Socrates [17] | - | 16 | - | 16 |
| Total | 1 | 272 | 230 | 503 |

| Dataset | Framerate | | | | Total |
| --- | --- | --- | --- | --- | --- |
| | 23-28 | 29-30 | 48 | 59-60 | |
| Youtube-8m [1] | 35 | 89 | - | - | 124 |
| Sports-1M [24] | 23 | 82 | - | - | 105 |
| Vision [43] | 23 | 73 | - | - | 96 |
| FloreView [4] | 2 | 155 | 1 | 4 | 162 |
| Socrates [17] | - | 16 | - | - | 16 |
| Total | 83 | 415 | 1 | 4 | 503 |

### 3.1 Data collection and generation

We collected a total of 503 real videos from five different datasets. YouTube-8M [1] and Sports-1M [24] are large-scale collections of YouTube videos originally used for action recognition. The remaining three datasets—Vision [43], FloreView [4], and Socrates [17]—consist of image and video data commonly used for camera source attribution [34].

These five datasets offer a diverse mix in terms of video content, resolution, frame rate, and media processing, ranging from shallow in-device compression to potential re-uploads on social media platforms. Additional details on the original video resolution and frame rate are provided in Table 1.

We selected five of the most recent open-source DM-based video generative techniques. SEINE [12] and DynamiCrafter [50] belong to the *frame, prompt-to-video* category, while RAVE [23], TokenFlow [18], and Text2Video-Zero [26] fall under *video, prompt-to-video*. Additional details on the selected generative techniques are provided in Sect. 3.2. Given the *prompt*-guidance needed by all generative pipelines, each video must be associated with a textual prompt which will be used during the denoising process.

While all the selected generative techniques rely on textual prompts, different DM generative pipelines require different types of prompts. *Frame, prompt-to-video* models, such as SEINE and DynamiCrafter, are typically used to re-enact video frames or create transitions between them. In contrast, *video, prompt-to-video* models are often employed for video editing, such as substituting or adding objects (as seen with RAVE and TokenFlow), or performing global edits (e.g., changing the season, altering weather conditions) as in Text2Video-Zero.

For each of the $N = 503$ original videos, we generated a corresponding prompt description by inputting the first frame into Blip2 [32], a VLM for image captioning. This process produced a set of *unaltered* prompts, denoted as $\mathcal{P}_u = \{p_{u_n} \mid n = 1, \ldots, N\}$, where each $p_{u_n}$ represents an individual prompt within $\mathcal{P}_u$. These prompts provide neutral descriptions of the videos to guide the generative process.

The set of *unaltered* prompts $\mathcal{P}_u$ were then manually modified to create a set of *altered* prompts, $\mathcal{P}_a$. These were used to generate videos similar to the original ones but with local modifications—such as edited objects or backgrounds—to be applied by the generative DM techniques [12, 18, 23, 26, 50].

Finally, we constructed a third set of *scene-editing* prompts, $\mathcal{P}_s$, describing specific global changes (i.e., season, sky, or weather alterations) to be applied to a video. These modifications were designed to affect the overall scene while ensuring that no elements were added, removed, or altered in isolation.

The two *frame, prompt-to-video* techniques, DynamiCrafter [50] and SEINE [12], use $\mathcal{P}_u$ during the video generation process. On the other hand, the first two *video, prompt-to-video* techniques, TokenFlow [18] and RAVE [23], use $\mathcal{P}_a$, while Text2Video-Zero [26] utilizes $\mathcal{P}_s$.

Through this approach, the DM-based videos in VideoDiffusion encompass a broad range of video-generation pipelines, including still image animation, transitions, local editing, and global editing.

The videos were generated by first extracting frames from the original videos. These frames were then resized to $640 \times 360$ pixels using bilinear interpolation and saved in PNG format. The DM-generated video frames were produced using the corresponding set of prompts ($\mathcal{P}_u, \mathcal{P}_a, \mathcal{P}_s$) required by each generative technique, and re-encoded as H.264 streams at three different compression rates, specified by CRF = $\{23, 30, 50\}$.

The final VideoDiffusion dataset contains 503 original videos and $503 \times 5 = 2515$ synthetically generated ones, all with resolution $640 \times 360$ pixels. Each video has four compression levels: uncompressed, CRF 23, CRF 30, and CRF 50; for a total of 2012 real and 10060 synthetic videos. Some examples are depicted in Fig. 2.

## 3.2 Generative techniques

*3.2.1 SEINE.* [12] is a *frame,prompt-to-video* technique based on LaVie [48], a *prompt-to-video* foundation model comprised of three cascaded video DMs. The first, LaVie-base, is a *prompt-to-video* model adapted from Stable Diffusion 1.4, incorporating Rotary Positional Encoding [48] for temporal attention and then jointly finetuned for image and video generation to avoid catastrophic forgetting [27]. SEINE extends LaVie-base with the addition of a mask module, used to selectively suppress the frame information during training, and thus making it learn how to generate from noise the masked in-between frames. This training objective provides SEINE with the ability to animate still images, extend short videos, and generate transitions between images by "filling the gap" between the selected input frames. During the video generation, SEINE generates transitions between temporally separated frames sampled from each original video.

*3.2.2 DynamiCrafter (DC).* [50] is a *frame,prompt-to-video* DM generative technique built upon VideoCrafter1 [10] (a DM trained for *prompt-to-video* and *frame, prompt-to-video* generation), and able to better preserve visual details by further injecting the original image during the denoising process.

The frame-injection at inference time is performed in two complementary phases. First, by concatenating the input frame to the random noise in input to the U-net to provide guidance for the local details during the denoising process. Then, by means of an additional cross-attention mechanism [50] on the input frame, which is able to provide a richer global context representation than using only the prompt, as commonly seen in *prompt-to-media* DM generators.

During the generation of our dataset, DynamiCrafter operates using only the first frame of each real video. Moreover, we removed the 16-frame generation limit from the original implementation [10]. The visual quality of the generated frames is akin to the other generative techniques used for VideoDiffusion and the short-scale time-consistency was deemed satisfactory.

*3.2.3 RAVE.* [23] is a *video,prompt-to-video* technique, which leverages pretrained *image,prompt-to-image* capable models, such as Stable Diffusion [38] for frame-level editing. This enable to preserve the global structure of the original video while providing a flexible video editing framework leveraging pre-trained DMs without requiring additional training or fine-tuning.

However, due to the strictly 2D nature of these models, it is necessary to enforce temporal consistency across subsequent frames in a model-agnostic manner during the denoising process [37, 49].

To this goal, RAVE introduces a novel noise-shuffling strategy that ensures greater temporal consistency in generated videos. This is achieved by arranging video frames into grids and processing them as single images. During the denoising process, at each timestep, the frames within each grid are randomly shuffled, promoting spatio-temporal interactions. Convolutional layers smooth the latent vectors to reduce flickering, while self-attention layers maintain consistent style and structure across frames [23].

The original implementation was modified to resize the grids to the common $640 \times 360$ pixels, while using the default Stable Diffusion 1.5 [38] backbone.

*3.2.4 TokenFlow (TF).* Similarly to RAVE, [18] is a *video,prompt-to-video* method used for video editing which leverages pretrained *image-to-image* capable networks for local editing, while enforcing the necessary time-consistency in a model-agnostic manner. TokenFlow, like RAVE, applies only local editing to video frames preserving the global structure of the original video.

To ensure temporal consistency, the preprocessing step of Tokenflow finds correspondences between the output features of the self-attention modules computed for each frame of the original video, and uses them in the editing process to linearly combine the features of adjacent frames.

For the generation of our dataset, the editing was carried out using the default Stable Diffusion 2.1 [38] backbone.

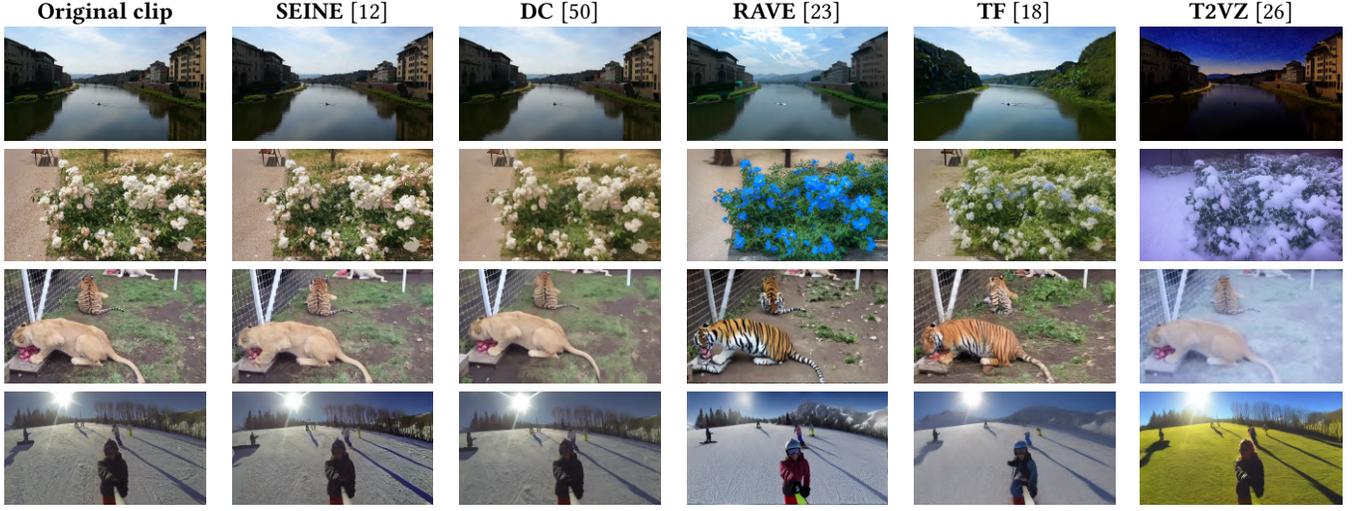

Figure 2: Original video frames and the corresponding generated ones using SEINE [12] and DynamiCrafter (DC) [50] (*frame,prompt-to-video* generative pipeline), RAVE [23] and TokenFlow (TF) [18] (*video, prompt-to-video* local editing), and Text2Video-Zero (T2VZ) [26] (*video, prompt-to-video* global editing).

*3.2.5 Text2Video-Zero (T2VZ).* [26] is a *video, prompt-to-video* generation technique that exploits pretrained Stable Diffusion [39] models, enhancing the temporal consistency by replacing self-attention layers with cross-frame attention ones.

By doing so, both visual information and overall structure of the first frame are preserved throughout the generation process, thus improving inter-frame consistency.

The selected *Video Instruct-Pix2Pix* [26] pipeline replaces the self-attention mechanism of the standard Instruct-Pix2Pix method [8], originally developed for instruction-based image-to-image editing.

The implementation was modified to pad the input frames to meet the requirement that both dimensions be divisible by 64; this padding was later removed from the resulting video frames.

## 4 PROPOSED METHOD

Synthetic video detection presents several challenges, as discussed in previous sections. One of the primary issues is the scarcity of video datasets, which hinders the development and training of robust detection models.

Moreover synthetic video generation involves multiple pipelines and, unlike static images, introduces additional complexities, such as temporal and inter-frame dependencies, and stronger compression artifacts, all of which complicate the detection process.

To address these challenges, we propose a novel technique that leverages high-level semantic temporal aware features, from the ViT originally proposed in [11] for video captioning.

As demonstrated in [15], ViT models provide high-level semantic features robust to data compression and post-processing , making them effective in distinguishing real media from fake ones. This robustness is particularly evident even with small training sets, thanks to ViTs outstanding few-shot learning capabilities [15].

Following the visual abstract of Fig. 1, the ViT we use, in contrast to [15] and [7], exploits both spatial and temporal information through a structured process involving four key components. First, a batch $B$ of $J$ video frames is resized to a resolution of $224 \times 224$ pixels and fed into a visual encoder, which generates frame representations $\mathbf{F} = [\mathbf{f}_1, \mathbf{f}_2, \ldots, \mathbf{f}_J]$, where $\mathbf{f}_j \in \mathbb{R}^{M_f \times D_f}$ consists of $M_f$ embeddings of dimension $D_f$ for the $j$-th frame. To incorporate temporal information, a positional encoding layer introduces positional embeddings $\mathbf{P} = [\mathbf{p}_1, \mathbf{p}_2, \ldots, \mathbf{p}_J]$, such that $\mathbf{P} \in \mathbb{R}^{J \times D_f}$, resulting in temporally aware features $\mathbf{F}' = \mathbf{F} + \mathbf{P}$. The enriched spatio-temporal features $\mathbf{F}'$ are then processed by a Video Query Transformer (Video Q-Former), which employs cross-attention layers to generate fixed-length video embeddings $\tilde{\mathbf{V}} \in \mathbb{R}^{T_v \times D_v}$, where $T_v$ represents the number of visual tokens and $D_v$ their dimension. Finally, a projection layer applies a linear transformation $\mathbf{W}_v \in \mathbb{R}^{D_v \times E}$, where $E$ denotes the original LLM's text embeddings' feature space dimension, to map $\tilde{\mathbf{V}}$ into a feature embedding $\tilde{\mathbf{V}}'$ of size $(T_v, E)$.

Our method takes the ViT from Fig. 1, freezes its weights, and builds a lightweight classifier on top, consisting of three linear layers that refine the feature representation. A first linear transformation $\mathbf{W}_1 \in \mathbb{R}^{T_v \times t}$ projects $\tilde{\mathbf{V}}'$ into an intermediate space of dimension $(t, E)$. Then, a second transformation $\mathbf{W}_2 \in \mathbb{R}^{E \times e}$ maps the feature representation to a lower-dimensional space $(t, e)$. Finally, the features are flatten and a third transformation $\mathbf{W}_3 \in \mathbb{R}^{E \times 1}$ further reduces the dimensionality to obtain the scalar $\boldsymbol{\omega}$. The final output $s(\boldsymbol{\omega})$ is produced using the learned prototypes [31, 41] of (1) where, with respect to the original implementation [31], we added $\text{sign}\left(\sum_{k=1}^{C}(\omega_k - c_k)\right)$ as this latter is beneficial for binary classification tasks.

$$s(\boldsymbol{\omega}) = \left[\text{sign}\left(\sum_{k=1}^{C}(\omega_k - c_k)\right) \cdot \left(\frac{\|\boldsymbol{\omega} - \mathbf{c}\|}{2\sigma_k^2}\right) + C \cdot \log \sigma_k\right]. \quad (1)$$

In (1), $C$ is the input feature dimension, and $\mathbf{c} = [c_1, \ldots, c_k] \in \mathbb{R}^C$ are centroids defining an isotropic Gaussian class-conditional distribution with standard deviations $\boldsymbol{\sigma} = [\sigma_1, \ldots, \sigma_k] \in \mathbb{R}^C$. Both **c** and $\boldsymbol{\sigma}$ are learnable parameters, with $\log \sigma_k > 0, \forall_k$ to make the distance function satisfies triangular inequality, a necessary condition to use learned prototypes as a metric [31].

Learned prototypes [31, 41] are typically used to detect out-of-distribution (OOD) samples in classification tasks. In our framework, we leverage them during training to infer the parameters **c** and $\boldsymbol{\sigma}$, which define a Gaussian class-conditional distribution for real videos. Consequently, all OOD videos are classified as fake. This design choice further enhances robustness against H.264 compression with CRF $\geq 23$.

During testing, our method samples 64 frames from a video dividing them into $B = 8$ uniformly sampled, time-continuous batches, each containing $J = 8$ frames. The final decision $\phi$ is obtained by aggregating the results of each batch into a single video-level embedding [45] and applying a sigmoid function $\sigma(x)$:

$$\phi = \sigma\left(\sum_{b=1}^{B} s(\omega_b)\right). \quad (2)$$

The training process was kept as simple as possible. Sequences of $J = 8$ consecutive frames were sampled from the same real and fake videos, forming $B = 5$ mini-batches used to compute the binary cross-entropy (BCE) loss, shown in equation (3):

$$\mathcal{L}_{\text{BCE}} = -\frac{1}{B} \sum_{b=1}^{B} \left[y_b \log(\hat{y}_b) + (1 - y_b) \log(1 - \hat{y}_b)\right], \quad (3)$$

where $y_b$ is the ground truth label (i.e., 1 or 0) for the $b$-th mini-batch, and $\hat{y}_b = \sigma(s(\boldsymbol{\omega}))$ is the predicted probability of the same mini-batch[2].

### 4.1 Training hyperparameters

The dataset is partitioned into training, validation, and test sets with a 70/10/20 splits. The training set consists of a total of 1408 videos, equally divided between real and fake samples. In details, first we sampled $M = 352$ real videos, then following [15], we selected $M = 352$ synthetic videos generated with TokenFlow [18], and $M = 352$ with DynamiCrafter [50], that shared the same unaltered prompt description $\mathcal{P}_u$ with the real videos.

Tokenflow and DynamiCrafter have been selected to ensure that all methods are exposed to both generative pipelines during training: *frame, prompt-to-video* and *video, prompt-to-video*. This condition will be crucial in Sect. 5.2 for assessing the real transfer-learning capabilities of all methods in the presence of an unknown generator and generative pipelines such as *prompt-to-video*.

The selected video frames are not subjected to any H.264 compression and real videos are sampled twice to balance the number of real and fake samples.

During training, we use the Adam optimizer with a learning rate of $1 \times 10^{-4}$, a learning rate decay to $\frac{1}{10}$ every 5 epochs without improvement, a minimum learning rate of $1 \times 10^{-7}$, and a maximum of 200 epochs. The number of centroids used for the learned prototypes is set to $C = 1$. Training typically converges after 40 epochs.

---
[2]Full-code is available at https://github.com/MMLab-unitn/VideoDiffusion-IHMMSec25

Data augmentation consists of mutually-exclusive Gaussian blur, using a kernel size of 15 and a sigma in the range (0,3), or JPEG compression with a quality factor in the range [65, 95], with a probability of 0.3.

Additionally, we apply Gaussian blur over multiple consecutive frames with a smaller kernel size of 3, occurring with a probability of 0.2. We also use vertical and horizontal flips, applied with a probability of 0.5, and random resized cropping to 300 × 300 pixels with a scaling factor in the range [0.9, 1.1] and ratio [0.75, 1.33], ensuring consistency in cropping position across all frames of the same video, with probability of 0.2.

## 5 EXPERIMENTAL RESULTS

We evaluate our method in terms of single-class True Positive Rate (TPR) and True Negative Rate (TNR), Area Under the Curve (AUC), and accuracy against five solutions from the SoA. The TPR reflects the detectors' ability to identify fake videos, while the TNR to detect real ones.

The first three SoA baselines are the CLIP-based architecture (CLIP-D) from [15] (with code available at [21]), adapted for video analysis via majority voting on individual frame results [7], AIGVDet [3], and MISLNet [45]. To these baselines, we added the ResNet-50 model from [13], hereafter referred to as R50-ND, which is known for its success in DM image detection and has demonstrated promising performance on fake videos, as shown in [45]. Additionally, we incorporated the ResNet3D from [16], initially implemented for video recognition tasks, adapting it for fake video detection. The ResNet3D classification head was modified by replacing the Average 3D Pooling with Adaptive Average Pooling and producing the final score through the learned prototypes [2]. The ResNet3D weights were first initialized using the pre-trained model for action recognition, then finetuned *end-to-end*, on our training set. During training, we use the same batch organization, data augmentation, and loss function (3) as for our proposed method, but with $B = 1$ and $J = 16$. Conversely, during testing, the batch size was set $B = 8$ and $J = 16$. In summary, we compare our solution with three methods based on frame-based predictions (i.e., MISLNet [45], R50-ND [13] and CLIP-D [15]), and two that depends by spatio-temporal features inferred from frame sequences (i.e., the ResNet3D and AIGVDet [3])

In all our training and testing, except where specified otherwise, we used the implementation provided by each method's original paper. Our only adaptations involved AIGVDet [3], where the crop size was reduced from 448 × 448 to 224 × 244 pixels due to the lower resolution (640 × 360) of VideoDiffusion's videos. Additionally, the maximum number of test frames processed by [13] and [21] was set to 80 frames, as was done for MISLNet [45]. During testing, all methods thresholds are fixed equals to 0.5 to simulate realistic scenario where no prior information is available to perform any calibration.

In our first two sets of experiments, we evaluate all methods in terms of single-class TPR and TNR, and AUC, first on VideoDiffusion (also used during training) and then on a disjoint dataset composed of videos generated by proprietary *prompt-to-video* techniques to assess their learning transferability performance. Next, we evaluated the few-shot learning capability of our method by replicating the experiments of [15], measuring the accuracy obtained using the entire training set $M = 352$, as well as reduced versions consisting of only $M = 10$ or $M = 100$ videos.

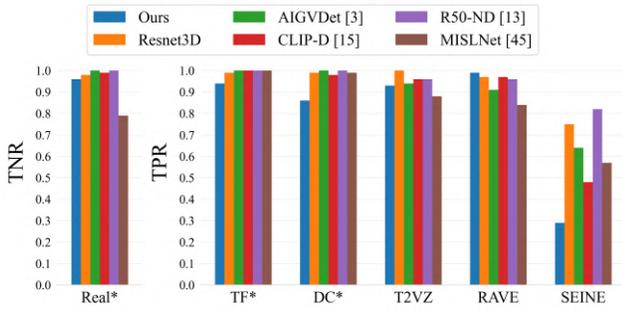

Figure 3: Results in terms of single-class TPR and TNR obtained with the proposed method and the selected baseline on uncompressed frames from real and fake videos. With * we refer to the classes of videos seen also during training. With TF, DC, and T2VZ we refer respectively to TokenFlow, DynamiCrafter and Text2Video-Zero.

All architectures were trained on the same training set of uncompressed video frames described in Sect. 4.1, and all experiments conducted on a server with the following specifications: NVIDIA GeForce RTX 3090, 14-core Intel(R) Core(TM) i9-10940X CPU @ 3.30GHz, and 256 GB RAM.

## 5.1 Results on VideoDiffusion

In this section, we evaluate the performance of our method and those of the SoA in terms of single-class TPR, TNR, and AUC on videos from the proposed dataset, VideoDiffusion. Additionally, we measure the respective performance degradation when tested on H.264 compressed video frames with CRF values of 23, 30, and 50.

In Fig. 3, we present the results obtained when testing on uncompressed video frames. From these initial results, we observe a clear positive trend for all methods that exploit both spatial-temporal features, specifically our method, the ResNet3D, and AIGVDet [3]. Comparable results are also obtained by MISLNet [45] and other frame-based video detectors, such as the R50-ND [13] and CLIP-D [15]. However, as we will show in the subsequent results, the performance degradation of these latter is significantly higher in presence of H.264 compression and for generative pipelines unseen during training.

A notable challenge for all tested methods is detecting SEINE-generated videos. This issue is particularly pronounced for models based on pretrained ViTs, such as ours and CLIP-D [15]. In contrast, fully retrained CNN architectures, including ResNet3D, MISLNet, and AIGVDet, adapt more effectively to SEINE-generated videos, achieving better recall.

This discrepancy is likely due to the training paradigm used for SEINE, in which missing frames are inferred from neighboring available ones. This process presumably helps SEINE modelling the latent space, leveraging the time-aware denoising network [48], to enhance semantic consistency by propagating information across multiple frames.

| CRF \ Detector | Ours | Resnet3D | AIGVDet | CLIP-D | R50-ND | MISLNet |
|---|---|---|---|---|---|---|
| Uncomp. | 0.88 | 0.96 | 0.95 | 0.94 | **0.97** | 0.83 |
| 23 | 0.79 | **0.87** | 0.73 | 0.58 | 0.61 | 0.62 |
| 30 | 0.70 | **0.72** | 0.55 | 0.56 | 0.52 | 0.52 |
| 50 | **0.67** | 0.57 | 0.59 | 0.57 | 0.51 | 0.50 |

Table 2: Average AUC scores obtained by each detector on VideoDiffusion. With "Uncomp." we refer to uncompressed video frames, stored as PNGs and not subjected to any H.264 video compression.

As a result, the semantic information from real frames used in generation may propagate in a way that evades detection in the high-level feature space captured by ViTs. This occurs even when visible artifacts, noticeable to the human eye, are present—artifacts that 2D and 3D CNN-based detectors capture more effectively.

In Fig. 4, we repeated the previous experiment on videos compressed with H.264 at CRF= 23, 30, and 50. As evident in Fig. 4, the proposed method, along with ResNet3D and AIGVDet [3] also relying on spatio-temporal features, demonstrates greater resilience to H.264 video compression, even when trained on uncompressed data. Additionally, while our method performs comparably to ResNet3D and AIGVDet [3] at CRF 23, its performance degradation is much lower for CRF=30 and 50. This is particularly true when comparing TNR and TPR obtained on real and generated videos, especially for generative techniques never seen during training (i.e., T2VZ and RAVE).

The superior adaptability of methods leveraging ViT's high-level semantic features can also be grasped by observing the results obtained with CLIP-D [15] for CRF 50. Indeed, CLIP-D is the only frame-based detector that performs comparably to the proposed solution, ResNet3D, and AIGVDet on both real and synthetic videos.

Finally, the results of Fig. 3 and Fig. 4 are confirmed by Table 2, where we report the AUCs obtained by each detector on the different compressed versions of *VideoDiffusion*. Specifically, detectors that are spatio-temporal aware, such as the proposed ones, ResNet3D, and AIGVDet, consistently exhibit less performance degradation than the other detectors.

## 5.2 Results on a disjoint dataset

Once we assessed the performance in terms of single-class TPR, TNR, and AUC, of our solution and those of the SoA on VideoDiffusion, we aimed to measure the learning transferability of all methods when tested on a disjoint dataset of videos generated by proprietary techniques. These settings are particularly challenging because neither the generators nor their generative pipeline (i.e. *prompt-to-video*) have been encountered during training by any of the detectors.

The dataset used in this section consists of 41 videos generated with SORA [9], 49 with LUMA AI [29], 74 using RunwayML [33], 250 using Hunyuan [28], 250 using CogVideo [51], and 101 full-resolution real videos from the VideoDiffusion test-set (i.e., no resizing to 640 × 360 was applied). Video resolutions in this dataset vary from 720 × 480 for CogVideo and Hunyuan to 1168 × 864 for LUMA AI, 1024 × 640 for RunwayML, up to 1980 × 1080 for SORA-generated videos and real ones. Each video created by these generative techniques is unique, meaning they were not generated using the same prompts. We show a few examples of this dataset in Fig. 5.

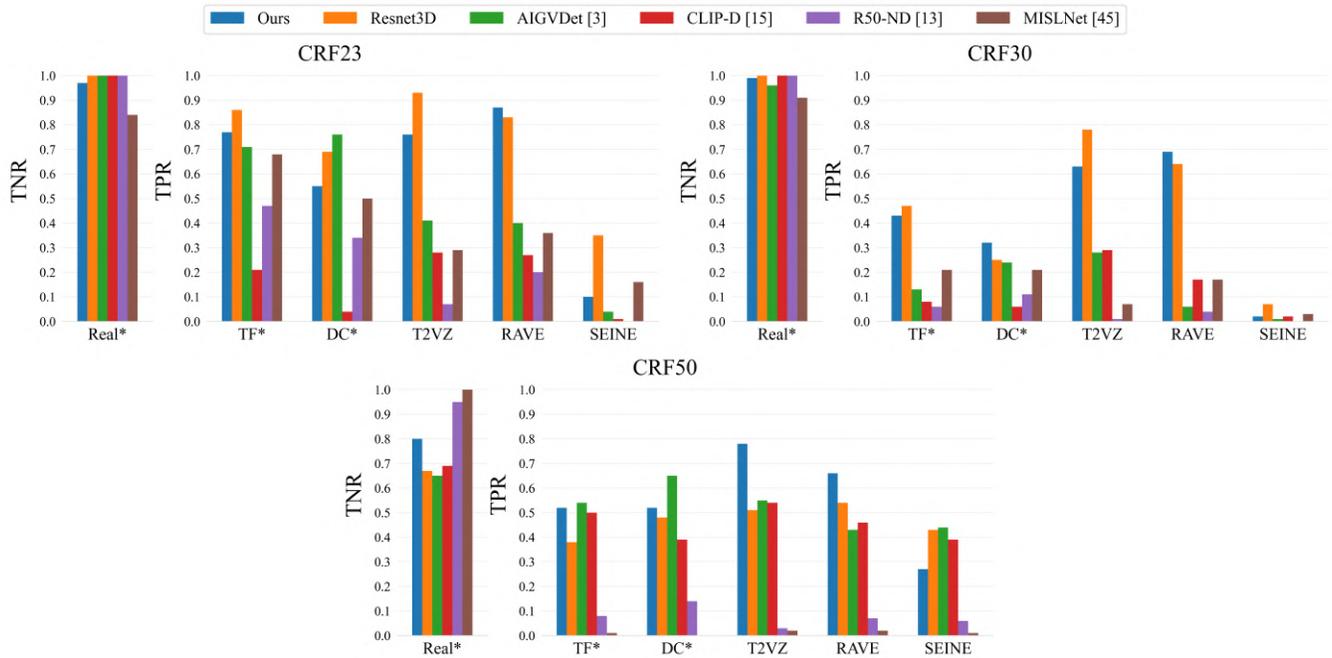

Figure 4: Results in terms of single-class TPR and TNR obtained with the proposed method and the selected baselines on compressed frames from real and fake videos. With * we refer to classes of videos seen during training but as uncompressed video frames.

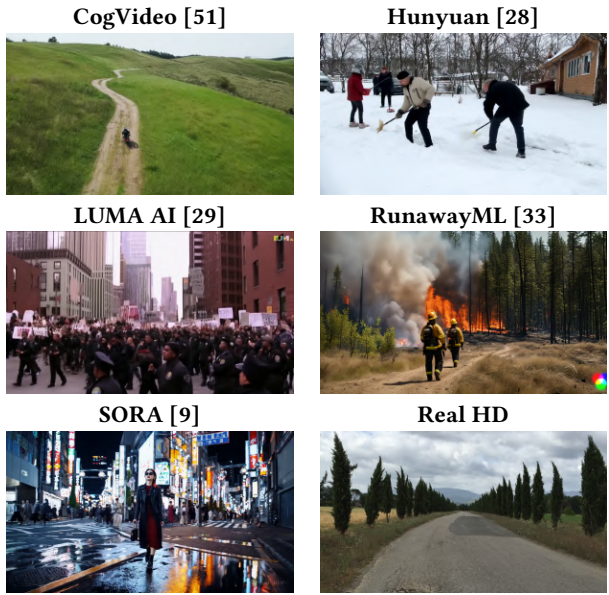

Figure 5: Disjoint dataset showcase, the videos were generated with several proprietary state-of-the-art *prompt-to-video* models. *Real HD* is the same collection from [1, 4, 17, 24, 43] described in Sec. 3.1, without the resize operation.

In Fig. 6, we show the results in terms of single-class TPR and TNR obtained on the dataset of proprietary generative techniques presented in this section. Notably, our approach, which leverages high-level temporally-aware semantic features, is the only one capable of successfully generalizing across all classes of real and fake videos obtained by previously unseen *prompt-to-video* generative pipeline.

AIGVDet [3], which also utilizes temporally-aware features extracted from dense optical flow estimation, is the second-best performing technique. Meanwhile, CLIP-D [15], another ViT-based model, confirms the strong generalization capabilities demonstrated in its original paper. This is because ViTs, including the one proposed for our solution, benefit from pre-training on a much larger and more diverse dataset, allowing them to capture higher-level semantic features. This improves generalization without requiring additional fine-tuning.

In contrast, methods relying on low-level features (e.g., R50-ND [13], MISLNet [45] and ResNet3D) exhibit weaker transferability, achieving high TNRs only for real videos and TPRs for fake ones generated with RunwayML. Indeed, we observed that while ResNet3D demonstrates remarkable detection abilities on VideoDiffusion (including compressed videos), it severely underperforms when tested on proprietary generators.

There are likely two main reasons for this: First, during the experiments in Sect. 5.1, although some generators were unknown,

| Detector Gen. Tech. | Ours | Resnet3D | AIGVDet | CLIP-D | R50-ND | MISLNet |
|---|---|---|---|---|---|---|
| SORA | **0.95** | 0.53 | 0.52 | 0.56 | 0.49 | 0.55 |
| LUMA AI | **0.96** | 0.66 | 0.88 | 0.68 | 0.57 | 0.47 |
| Hunyuan | **0.92** | 0.67 | 0.79 | 0.82 | 0.59 | 0.52 |
| CogVideo | **0.91** | 0.71 | 0.86 | 0.68 | 0.78 | 0.68 |
| RunwayML | 0.97 | 0.69 | 0.89 | 0.95 | **0.98** | 0.57 |

**Table 3: Average AUC scores obtained by each detector on the disjoint dataset of fully generated videos.**

similar generative pipelines (i.e. *frame, prompt-to-video* and *video, prompt-to-video*) had been encountered during the original training; Second, we found that the ResNet3D [16] was able to perform well in fake video detection tasks only when extensively finetuned *end-to-end*. As a result, finetuning the ResNet3D on a limited subset of synthetic videos may have hindered the model's generalization capabilities, aligning with the findings of [47].

To conclude, the results of Fig. 6 are further confirmed by the AUCs reported in Table 3. Indeed, across all these results, better TPR, TNR, AUC, and learning transferability are achieved by leveraging temporally-aware higher-level semantic features, as demonstrated by the solution proposed in this paper. Alternatively, methods incorporating either spatio-temporal or high-level semantic features, as for AIGVDet [3] and CLIP-D [15], also show good performance.

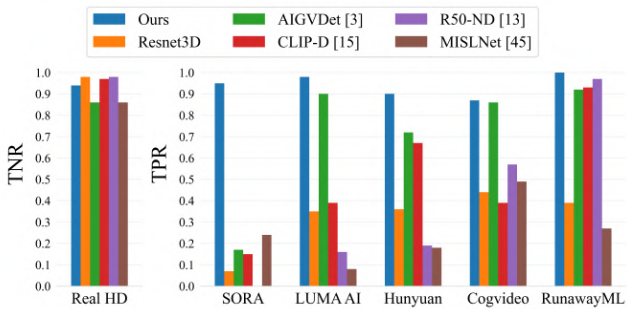

**Figure 6: Results in terms of single-class TPR and TNR obtained with the proposed method and the selected baseline on the disjoint dataset of videos generated by proprietary techniques.**

### 5.3 Few-shot learning capabilities

Inspired by [15], in this section we investigate few-shot learning proprieties of our proposed solution. We trained our model using training sets with $M \in \{10, 100, All\}$ (where All= 352 is the entire training-set) real and fake videos, and following the same data preparation described in Sect. 4.1. Training video frames were not subject to H.264 compression and were selected from the original training partition of TokenFlow [18] and DynamiCrafter [50]. Moreover, as we have already measured in Sect. 5.1 and 5.2 the detection performance of the proposed solution and the baselines on each fake and real video class, we now are interested in evaluating the accuracy degradation of our method as a function of M on VideoDiffusion and the dataset used in Sect. 5.2.

Fig. 7 demonstrates the few-shot capability of the proposed method in learning how to distinguish real from synthetic videos, as well as a clear trend between the training set size and the method's accuracy. Interestingly, we observe that with just $M = 100$ videos (less than half the size of the original training set proposed in Sect. 4.1) the accuracy on both real and generated videos, subjected to different levels of H.264 compression, remains comparable to that obtained using the entire training set. Moreover, although M= 10 represents an extreme case of data scarcity, we demonstrate that accuracies approaching, and sometimes bigger than, 70% are still achievable in this scenario on uncompressed or H.264 compressed with CRF 23 video frames.

In real-world scenarios, where new fake video generators are developed every year, the striking few-shot learning capabilities of ViT-based detectors we observed are essential. In fact, while in the long run, continuously producing large and diverse datasets may become unfeasible, Fig. 7 illustrates our solution's ability to generalize both to unknown generators and H.264 compression, even when trained with M≤ 100.

## 6 CONCLUSIONS

In this paper, we proposed an innovative solution for synthetic video detection based on high-level, temporal aware semantic features extracted using the ViT introduced in [11]. Additionally, we presented VideoDiffusion, a novel, large, and diverse dataset comprising over ten thousand videos generated by five open-source, DM-based video generation techniques.

We evaluated our solution against two frame-based CNNs for fake video detection [13, 45], a ViT-based approach [15] that operates frame-wise, and two video-based detectors, one based on the ResNet3D of [16], while the other [3] leveraging spatio-temporal features extracted at RGB and dense optical flow level.

Our experiments demonstrated several advantages of our approach. In the first evaluation on VideoDiffusion, where all methods were trained on the same subset of this dataset, our method exhibited superior resilience to H.264 video compression, particularly for CRF $\geq$ 30.

When tested on a disjoint dataset consisting of proprietary methods utilizing unknown generative pipelines, our approach was the only one capable of effectively generalize under these unseen test conditions. The superior performance of our method stems from its integration of both high-level semantic and spatio-temporal features, as evidenced by the second- and third-best-performing models [3, 15].

Furthermore, we demonstrated the promising few-shot learning capabilities of our solution on VideoDiffusion as well as on proprietary generative techniques.

As our experimental results suggest, in real-world scenarios where diffusion-based generators improve exponentially every year (making fake video generation increasingly difficult to regulate) the few-shot learning capabilities of ViT-based detectors are crucial. This property is particularly relevant for various applications, including efficient and reliable data labeling required by the EU Digital AI Act [36], continuous updates to detect emerging generative techniques, and *in-the-wild* multimedia forensics, especially for fake videos shared on social media [6].

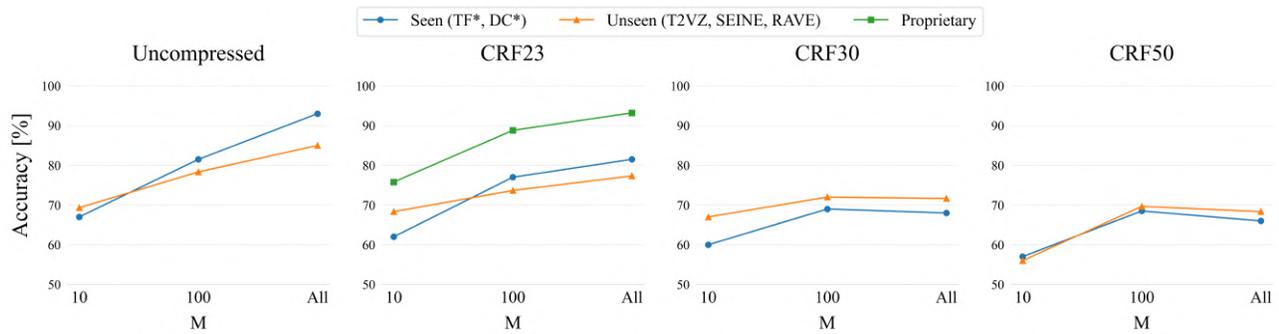

Figure 7: Accuracy of the proposed solution when trained respectively using M ∈ {10, 100, All}. The blue trends are the performance on generative techniques seen during training, while the orange ones on unseen generative techniques. The green trends show the accuracy obtained on the dataset of proprietary techniques.

Building on these observations, future work will focus on providing additional insights into the explainability of our solutions by verifying the coherence between DM-generated video residuals [13] as well as their feature space representations. In addition, we are interested in applying our method to fake and real videos shared on social networks, where data processing and compression significantly degrade the performance of fake media detectors [6]. Additionally, we aim to investigate whether the generalization and few-shot learning capabilities of our approach can help mitigating catastrophic forgetting when analyzing older deepfakes [42] or cheapfakes [46].

## ACKNOWLEDGMENTS

This work was partially supported by the European Union under the Italian National Recovery and Resilience Plan (NRRP) of NextGenerationEU (PE00000014 - program "SERICS"), the NGI Sargasso project DeepShield (101092887), and project Deepfake Detection 2.0 funded by Fondazione VRT.